\documentclass[sigconf]{acmart}
\AtBeginDocument{%
  \providecommand\BibTeX{{%
    \normalfont B\kern-0.5em{\scshape i\kern-0.25em b}\kern-0.8em\TeX}}}

\copyrightyear{2023}
\acmYear{2023}
\setcopyright{rightsretained}
\acmConference[WWW '23]{Proceedings of the ACM Web Conference 2023}{May 1--5, 2023}{Austin, TX, USA}
\acmBooktitle{Proceedings of the ACM Web Conference 2023 (WWW '23), May 1--5, 2023, Austin, TX, USA}
\acmDOI{10.1145/3543507.3583227}
\acmISBN{978-1-4503-9416-1/23/04}

\usepackage{subcaption}
\usepackage{colortbl}
\definecolor{zhz_gray}{rgb}{0.8,0.8,0.8}

\begin{document}

\title{TFE-GNN: A Temporal Fusion Encoder Using Graph Neural Networks for Fine-grained Encrypted Traffic Classification}

\author{Haozhen Zhang}
\affiliation{%
  \institution{Shenzhen International Graduate School, Tsinghua University}
  \country{China}
  }
\email{zhang-hz21@mails.tsinghua.edu.cn}

\author{Le Yu}
\affiliation{%
  \institution{Department of Computing, The Hong Kong Polytechnic University}
  \city{Hong Kong}
  \country{China}
  }
\email{yulele08@gmail.com}

\author{Xi Xiao}
\authornote{Corresponding authors.}
\affiliation{%
  \institution{Shenzhen International Graduate School, Tsinghua University}
  \country{China}
  }
\email{xiaox@sz.tsinghua.edu.cn}

\author{Qing Li}
\authornotemark[1]
\affiliation{%
  \institution{Peng Cheng Laboratory}
  \country{China}
  }
\email{liq@pcl.ac.cn}

\author{Francesco Mercaldo}
\affiliation{%
  \institution{University of Molise}
  \institution{IIT-CNR}
  \country{Italy}
  }
\email{francesco.mercaldo@unimol.it}

\author{Xiapu Luo}
\affiliation{%
  \institution{Department of Computing, The Hong Kong Polytechnic University}
  \city{Hong Kong}
  \country{China}
  }
\email{csxluo@comp.polyu.edu.hk}

\author{Qixu Liu}
\affiliation{%
  \institution{Institute of Information Engineering, Chinese Academy of Sciences}
  \country{China}
  }
\email{liuqixu@iie.ac.cn}

\begin{abstract}

Encrypted traffic classification is receiving widespread attention from researchers and industrial companies. However, the existing methods only extract flow-level 
features, failing to handle short flows because of unreliable statistical properties, or 
treat the header and payload equally, failing to mine the potential correlation between bytes. 
Therefore, in this paper, we propose a byte-level traffic graph construction approach based on point-wise mutual information (PMI), and a model named 
\textbf{T}emporal \textbf{F}usion \textbf{E}ncoder using \textbf{G}raph \textbf{N}eural \textbf{N}etworks (\textbf{TFE-GNN}) for feature extraction. 
In particular, we design a dual embedding layer, a GNN-based traffic graph encoder as well as a cross-gated feature fusion mechanism, which can first embed the header and payload bytes separately and then fuses them together to obtain a stronger feature representation. 
The experimental results on two real datasets demonstrate that TFE-GNN outperforms multiple state-of-the-art methods in fine-grained encrypted traffic classification tasks.

\end{abstract}



\maketitle

\section{Introduction}
\label{sec:intro}

To protect user privacy and anonymity, various encryption techniques are used to encrypt the transmission of network traffic~\cite{Encryption}. Although Internet security is improved for a regular user, encryption technologies also provide a convenient disguise for some malicious attackers.  
Moreover, some privacy-enhanced tools like VPN and Tor~\cite{VPNTor} may be utilized to achieve illegal network transactions, such as weapon trading and drug sales, where it is difficult to trace the traffic source~\cite{Threats}. 
Traditional data packet inspection (DPI) methods concentrate on mining the potential patterns or keywords in data packets, which is time-consuming and loses its accuracy when facing encrypted traffic~\cite{EncrySurvey}. 
Consequently, how to effectively represent encrypted network traffic for more accurate detection and identification is a significant challenge. 

To solve the above problems, many approaches have been proposed. 
The earliest port-based works are no longer effective due to the application of dynamic ports. Subsequently, a series of statistic-based methods emerged~\cite{GRAIN, KFP, AppScanner, ETC-PS, SAM}, which rely on statistical features from traffic flows (e.g., mean of packet length). 
Then, a machine learning classifier (e.g., random forest) is adopted to get the final prediction results. 
Unfortunately, these methods need hand-crafted feature engineering and may fail due to the unreliable/unstable flow-level statistical information in some cases~\cite{SFIM}. 
Most statistical features of relatively short flows have higher deviations compared with long flows. 
For example, the flow length generally obeys the long-tailed distribution~\cite{LongTail}, implying the universal existence of unreliable statistical features. Therefore, we use packet bytes instead of those statistical features.

Recently, graph neural networks (GNNs)~\cite{GCN} have been widely used in lots of applications of processing unstructured data. Due to the powerful expressiveness, GNNs can recognize specific topological patterns implied in graphs so that we can classify each graph with a predicted label. 
For the traffic classification task, most current GNN-based methods~\cite{GraphDApp, MAppGraph, MEMG, GAP-WF, ECD-GNN} construct graphs according to the correlation between packets, which actually is another usage form of statistical features and also suffers from the issue mentioned above. 
While the others do utilize packet bytes but have two major flaws: \textbf{1) Mix usage of the header and payload.} Existing methods simply treat the header and payload of a packet equally but ignore the difference in meaning between them. \textbf{2) Inadequate utilization of raw bytes.} Although the packet bytes are utilized, most methods regard packets as nodes and just take their raw bytes as node features, which does not make the most of them~\cite{ECD-GNN}.

Based on the above observations, in this paper, we propose a byte-level traffic graph construction approach based on point-wise mutual information (PMI) and a novel model named \textbf{T}emporal \textbf{F}usion \textbf{E}ncoder using \textbf{G}raph \textbf{N}eural \textbf{N}etworks (\textbf{TFE-GNN}) for encrypted traffic classification. 
The byte-level traffic graphs are constructed by mining the correlation between bytes and served as inputs for TFE-GNN. 
TFE-GNN consists of three major sub-modules (i.e., dual embedding, traffic graph encoder, and cross-gated feature fusion mechanism). 
The dual embedding treats the header and payload of a packet separately and embeds them using two independent embedding layers. 
As for the traffic graph encoder which consists of multilayer GNNs, it encodes each graph into a high-dimensional graph vector. 
Finally, we use the cross-gated feature fusion mechanism to integrate header graph vectors and payload graph vectors, obtaining an overall representation vector of a packet. 
For end-to-end training, we employ a time series model to get final prediction results for downstream tasks.
In the experiment section, we adopt a self-collected WWT dataset (including the data from WeChat, WhatsApp and Telegram) as well as the public ISCX dataset to compare TFE-GNN with more than a dozen baselines. The experimental results show that TFE-GNN surpasses almost all the baselines and comprehensively achieves the most excellent performance on the adopted datasets (e.g., 10.82\% ↑ on the Telegram dataset, 4.58\% ↑ on the ISCX-Tor dataset).

In summary, the main contributions of this paper include:
\begin{itemize}
\item We \textit{first} construct the byte-level traffic graph by converting a sequence of packet bytes into a graph, supporting traffic classification from a different perspective.
\item We propose TFE-GNN, which treats the packet header and payload separately and encodes each byte-level traffic graph into an overall representation vector for each packet. Thus, TFE-GNN utilizes a packet-level representation vector rather than a flow-level one. 
\item To evaluate the performance of the proposed TFE-GNN, we compare it with several existing methods on the self-collected WWT dataset and public ISCX dataset~\cite{ISCX, ISCX-Tor}.
The result shows that, for user behaviour classification, TFE-GNN outperforms these methods in effectiveness.
\end{itemize}

\section{Preliminaries}
\label{sec:preliminaries}

\subsection{Notations}
\label{sec:preliminaries_notation}

In this paper, a graph is denoted by $\mathcal{G}=\{\mathcal{V}, \mathcal{E}, \mathrm{X}\}$, where $\mathcal{V}$ is the node set, $\mathcal{E}$ is the edge set, and $\mathrm{X} \in \mathbb{R}^{|\mathcal{V}| \times d_{X}}$ is the initial feature matrix of nodes whereby the initial feature of node $v$ can be represented by $x_v$. 
We use $\mathcal{A} \in\{0,1\}^{|\mathcal{V}| \times |\mathcal{V}|}$ to represent the adjacency matrix of $\mathcal{G}$, which satisfies that the entry $(i, j)$ of $\mathcal{A}$, i.e., $a_{ij}$, equals 1 if there is an edge between nodes $i$ and $j$, otherwise it is 0.
We use $N(v)$ to represent the neighborhood of node $v$.
Moreover, we use $d_{l}$ to represent the embedding dimension in the $l$-th layer.

For brevity and convenience, we extend the concept of traffic flows by introducing time-induced \textbf{Traffic Segments (TS)}, which are collectively referred to as traffic samples in the rest of the paper. 
\begin{equation}
\emph{TS} = [P_{t_1}, P_{t_2}, \cdots, P_{t_n}], \ t_1 \leq t_2 \leq \cdots \leq t_n
\end{equation}
where $P_{t_i}$ denotes a single packet with its time stamp $t_i$, $n$ is the sequence length of a traffic segment, $t_1, t_n$ are the start and end times of a traffic segment, respectively. From the definition above, the traffic segment has a broader scope than the traffic flow, i.e., each traffic flow can be seen as a traffic segment, but the reverse does not necessarily hold. In this way, we can directly take traffic segments as training samples and do inference using either traffic flows or traffic segments, which helps to improve flexibility and unleash the expressiveness of an end-to-end model.

\subsection{Encrypted Traffic Classification}

The encrypted traffic classification task aims to differentiate the traffic generated from various sources (e.g., applications, web pages or services) by using the information of traffic packets captured by professional software or programs. 
In this paper, we concentrate on in-app user behaviour classification which differentiates fine-grained user actions such as sending texts and sending pictures.

Assume that there are $M$ training samples and $N$ categories in total, let the $i$-th traffic sample be a sequence $s_i = [bs_{1}^{i}, bs_{2}^{i}, \cdots, bs_{n}^{i}]$, where $n$ is the sequence length and $bs_j^{i}$ is the $j$-th byte sequence of the $i$-th traffic sample denoted by $bs_j^{i} = [b_1^{ij}, b_2^{ij}, \cdots, b_m^{ij}]$ where $m$ is the byte sequence length and $b_k^{ij}$ denotes the $k$-th byte value in the $j$-th byte sequence of the $i$-th traffic sample. According to the definition above, the (segment-level) encrypted classification task can be described formally as predicting the category $C_s$ of an unseen test sample $s_i$ with a designed and well-trained end-to-end model $F(s_i)$ on $M$ training samples, where $C_s = 0, 1, \cdots, N-1$.

\subsection{Message Passing Graph Neural Networks}
\label{sec:preliminaries_gnn}

Graph Neural Networks (GNNs)~\cite{GCN} are powerful models for handling unstructured data. 
With the application of the message passing paradigm (MP)~\cite{MP} to GNNs (MP-GNNs), the node embedding vectors can be updated iteratively by integrating nodes' embedding vectors in neighborhood through a specific aggregation strategy. Generally, the $l$-th layer MP-GNNs can be formalized as two procedures (i.e., the message computation and aggregation):
\begin{equation}
\begin{aligned}
\mathbf{m}_{u}^{(l)} &= \mathrm{MSG}^{(l)}\left(\mathbf{h}_{u}^{(l-1)}; \theta_m^l \right) \\
\mathbf{h}_{v}^{(l)} &= \mathrm{AGG}^{(l)}\left(\mathbf{h}_{v}^{(l-1)}, \left\{\mathbf{m}_{u}^{(l)}, u \in N(v)\right\}; \theta_a^l \right)
\end{aligned}
\end{equation}
where $\mathbf{h}_{u}^{(l)}, \mathbf{h}_{v}^{(l)} \in \mathbb{R}^{d_{l}}$ are the embedding vectors of nodes $u$ and $v$ in layer $l$.
$\mathbf{m}_{u}^{(l)}$ is the computed message from node $u$ in layer $l$. 
$\mathrm{MSG}^{(l)}(\cdot)$ is a message computation function parameterized by $\theta_m^l$ and 
$\mathrm{AGG}^{(l)}(\cdot)$ is a message aggregation function parameterized by $\theta_a^l$ in layer $l$. 
Notably, 
$\theta_m^l$ is optional and the inputs of MP-GNNs are given by initial node feature vectors (i.e., $\mathbf{h}_{v}^{(0)} = \mathbf{x}_v$).

Due to the high scalability of our proposed model, various GNN architectures can be 
easily adapted. 
Section \ref{sec:method_encoder} discusses the concrete choice of message aggregation strategies and our designed GNN architecture according to the design space of GNNs~\cite{GCN_design_space}.

\section{Methodology}
\label{sec:method}

\begin{figure*}[t]
	\centering
	\includegraphics[width=0.95\linewidth]{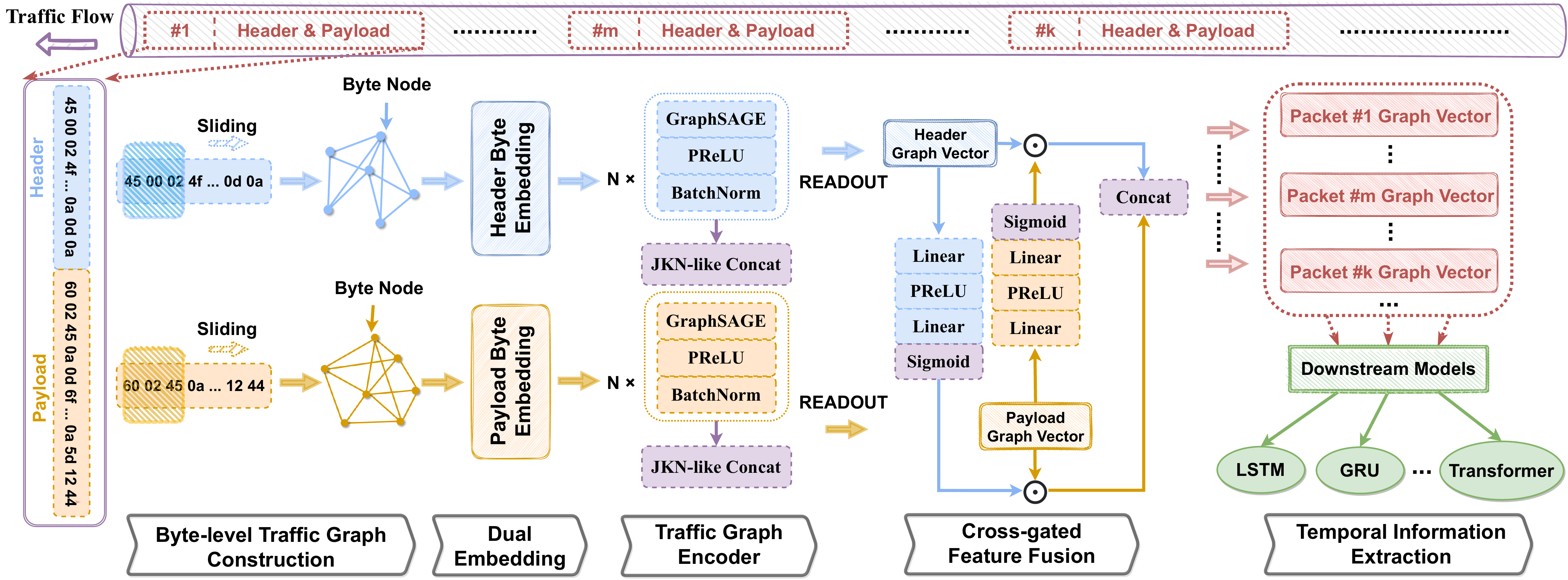}
	\caption{TFE-GNN Model Architecture}
	\label{model_figure}
        \Description[TFE-GNN Model Architecture]{TFE-GNN consists of several modules (Byte-level Traffic Graph Construction, Dual Embedding, Traffic Graph Encoder, Cross-gated Feature Fusion, and Temporal Information Extraction).}
\end{figure*}

\subsection{Byte-level Traffic Graph Construction}
\label{sec:method_graph_construction}
We attempt to convert a sequence of bytes into a graph $\mathcal{G} = \{\mathcal{V}, \mathcal{E}, \mathrm{X}\}$ by mining the potential correlation between bytes, where each element in $\mathcal{V}$ denotes a byte (i.e., a byte corresponds to a node in $\mathcal{G}$). 
Note that all the bytes with the identical value share the same nodes so that there are no more than 256 nodes in $\mathcal{G}$, which ensures a relatively small scale of traffic graphs. 

\textbf{Correlation representation between bytes.} For edges, we can easily connect all bytes chronologically, which means creating an edge from byte $i$ to $j$ if byte $i$ comes before byte $j$ in a byte sequence.
But we do not adopt this method since it will lead to a very dense graph and the topological structure will lack distinguishability. 
Therefore, inspired by~\cite{BLJAN} which uses cosine similarity to measure the correlation between two bytes, we adopt point-wise mutual information (PMI)~\cite{PMI}, which is a prevalent measure for word association computation in natural language processing (NLP), to model the correlation between two bytes. In this paper, we represent the PMI value of bytes $i$ and $j$ as $\operatorname{PMI}(i, j)$.

\textbf{Edge creation.} The PMI value makes a comprehensive measurement of two co-occurrence bytes from the perspective of semantic associativity of bytes. 
We utilize it to create an edge between two bytes. 
A positive PMI value implies a high semantic correlation of bytes while a zero or a negative one implies little or no semantic correlation of bytes. Consequently, we only create an edge between two bytes whose PMI value is positive. 

\textbf{Graph construction.} Below, we give the formal description of edges through the entries of adjacency matrix $\mathcal{A}$ of nodes $i$ and $j$:
\begin{equation}
a_{ij}= \begin{cases}1, & \operatorname{PMI}(i, j) > 0 \\ 0, & \text{Otherwise}\end{cases}
\end{equation}
The initial features of each node in graph $\mathcal{G}$ are given by the corresponding byte value, which ranges from 0 to 255. Notably, since $\operatorname{PMI}(i, j)=\operatorname{PMI}(j, i)$, the byte-level traffic graphs are undirected.

\subsection{Dual Embedding}
\label{sec:method_embedding}
The byte value is commonly utilized to serve as initial features for further vector embedding. 
Two bytes with different values correspond to two distinct embedding vectors. 
However, the meaning of a byte varies not only with the byte value itself, but also with the part of the byte sequence in which it is located. 
In other words, the representation meaning of two bytes with the identical value within the header and payload of a packet respectively may be completely different.
The reason is that the payload carries the transmission contents of a packet while the header is the first part of a packet that describes its contents. 
If we make two bytes with the identical value in the header and payload correspond to a same embedding vector, it is difficult for a model to converge to the optimum on these embedding parameters because of the obfuscated meaning.

For the rationale mentioned above, we treat the header and payload of a packet separately and construct byte-level traffic graphs for the two parts, respectively (i.e., byte-level traffic header graphs and byte-level traffic payload graphs). 
We adopt dual embedding with two embedding layers that do not share parameters to embed initial byte value features into high-dimensional embedding vectors for the two kinds of graphs, respectively.

\textbf{Dual embedding layer.} 
Assume that $d_0$ denotes the embedding dimension and $K$ is the number of embedding elements (i.e., byte value). 
The dual embedding matrices, which consist of two embedding matrices, can be viewed as $E_{header} \in \mathbb{R}^{K \times d_0}$ and $E_{payload} \in \mathbb{R}^{K \times d_0}$, where each row-wise entry represents the embedding vector of each byte value.

\subsection{Traffic Graph Encoder with Cross-gated Feature Fusion}
\label{sec:method_encoder}

Since we construct byte-level traffic graphs based on the header and payload of packets, respectively, the following modules of TFE-GNN in this section are also dual, do not share parameters (architecture is the same) and can process in parallel. 

\textbf{Traffic graph encoder.} To encode each traffic graph into a graph feature vector, we elaborately design a traffic graph encoder using stacked GraphSAGE~\cite{GraphSAGE}, which is a powerful graph neural network. For every node $v$ in graph $\mathcal{G}$, GraphSAGE computes the message from each neighboring node $u \in N(v)$ by normalizing its embedding vector using the degree of node $v$. Then, GraphSAGE computes the overall message of all neighboring nodes $N(v)$ through element-wise mean operation and aggregates the overall message as well as the embedding vector of node $v$ through concatenation operation. Finally, a nonlinear transformation is applied to the embedding vector of node $v$, finishing the forward procedure of one GraphSAGE layer. Formally, the message computation and aggregation of GraphSAGE can be described by: 
\begin{equation}
\begin{aligned}
\mathbf{m}_{N(v)}^{(l)} &= \sum_{u \in N(v)} \frac{\mathbf{h}_{u}^{(l-1)}}{|N(v)|} \\
\mathbf{h}_{v}^{(l)} &= \sigma\left(\mathbf{w}^{(l)} \cdot \operatorname{CONCAT}\left(\mathbf{h}_{v}^{(l-1)}, \mathbf{m}_{N(v)}^{(l)}\right)\right)
\end{aligned}
\end{equation}
where $|N(v)|$ is the neighbor number of node $v$, $\mathbf{w}^{(l)} \in \mathbb{R}^{d_{l-1} \times d_l}$ is the parameter in layer $l$, $\operatorname{CONCAT}(\cdot)$ denotes the concatenation operation and $\sigma(\cdot)$ denotes the activation function. Specially, we employ parametric ReLU (PReLU)~\cite{PReLU} as an activation function. 
PReLU scales each negative element value by a factor, which not only plays the effect of nonlinear transformation but also plays a role similar to that of the attention mechanism by different scale factors for each channel in the negative axis. Lastly, we normalize the updated feature vector $\mathbf{h}_{v}^{(l)}$ by batch normalization (BN)~\cite{BN}.

Due to the over-smoothing issue~\cite{Over-smoothing} in the deep GNN model, we only stack GraphSAGE up to $4$ layers and concatenate the output feature vectors of each layer for each node $v$ to alleviate this problem, which is similar to Jumping Knowledge Network (JKN)~\cite{JKN}:
\begin{equation}
\mathbf{h}_{v}^{\text{final}} = \operatorname{CONCAT}(\mathbf{h}_{v}^{(1)}, \mathbf{h}_{v}^{(2)}, \mathbf{h}_{v}^{(3)}, \mathbf{h}_{v}^{(4)})
\end{equation}
where $\mathbf{h}_{v}^{\text{final}}$ is the final feature vector of node $v$. Finally, we apply mean pooling on all nodes to get a graph feature vector $\mathbf{g}$:
\begin{equation}
\mathbf{g} = \frac{\mathbf{h}_{1}^{\text{final}} \oplus \ldots \oplus \mathbf{h}_{|\mathcal{V}|}^{\text{final}}}{|\mathcal{V}|}
\end{equation}
where $\oplus$ denotes element-wise addition. For simplicity, we use $\mathbf{g}_h$ and $\mathbf{g}_p$ to represent graph feature vectors extracted from traffic header graphs and traffic payload graphs, respectively. 

\textbf{Cross-gated feature fusion.} Since we extract features from traffic header graphs and traffic payload graphs respectively mentioned in Section \ref{sec:method_embedding}, we aim to create a reasonable relationship between $\mathbf{g}_h$ and $\mathbf{g}_p$ to get an overall representation of packet bytes. To this end, we carefully design a feature fusion mechanism named cross-gated feature fusion, to fuse $\mathbf{g}_h$ and $\mathbf{g}_p$ into a final encoded feature vector for each packet.

As shown in Figure \ref{model_figure}, we adopt two filters, each of which consists of two linear layers with a PReLU activation function between them.
First the two filters, which do not share parameters, are applied to $\mathbf{g}_h$ and $\mathbf{g}_p$, respectively and then an element-wise sigmoid function is used to scale each element to $[0, 1]$. 
We consider the scaled vectors as gated vectors ($\mathbf{s}_h$ and $\mathbf{s}_p$ for the header and the payload) and use them to crosswise filter the corresponding $\mathbf{g}_h$ and $\mathbf{g}_p$. 
Such a mechanism allows the model to filter out unimportant information and reserve the significant one for the two feature vectors. 
As the first part of the packet, the header describes its important features. Thus, it is reasonable to use header gated vector $\mathbf{s}_h$ to filter payload graph feature vector $\mathbf{g}_p$ and conversely use payload gated vector $\mathbf{s}_p$ to filter header graph feature vector $\mathbf{g}_h$. 

The cross-gated feature fusion can be formally represented by:
\begin{equation}
\mathbf{s}_h = \operatorname{Sigmoid}(\mathbf{w}_{h2}^T\operatorname{PReLU}(\mathbf{w}_{h1}^T \mathbf{g}_h + \mathbf{b}_{h1}) + \mathbf{b}_{h2})
\end{equation}
\begin{equation}
\mathbf{s}_p = \operatorname{Sigmoid}(\mathbf{w}_{p2}^T\operatorname{PReLU}(\mathbf{w}_{p1}^T \mathbf{g}_p + \mathbf{b}_{p1}) + \mathbf{b}_{p2})
\end{equation}
\begin{equation}
\mathbf{z} = \operatorname{CONCAT}(\mathbf{s}_h \odot \mathbf{g}_p, \mathbf{s}_p \odot \mathbf{g}_h)
\end{equation}
where $\mathbf{w}_{h1}, \mathbf{w}_{h2}, \mathbf{w}_{p1}, \mathbf{w}_{p2}  \in \mathbb{R}^{d_g \times d_g}$ and $\mathbf{b}_{h1}, \mathbf{b}_{h2}, \mathbf{b}_{p1}, \mathbf{b}_{p2}  \in \mathbb{R}^{d_g}$ are the weights and biases of linear layers. The symbol $\odot$ denotes element-wise product and $\mathbf{z}$ is the overall representation vector of the packet bytes, which can be used for the downstream tasks.

\subsection{End-to-End Training on Downstream Tasks}
\label{sec:method_temporal}
Based on the overall representation vector $\mathbf{z}$ for each packet, a packet-level or a segment-level classification task can be easily solved using a downstream classifier. We primarily focus on the segment-level task in this paper. 

\textbf{Temporal information extraction.} 
Since we have already encoded raw bytes of each packet in a traffic segment into a representation vector $\mathbf{z}$, the segment-level classification task can be considered as a time series prediction task. Here, we just adopt long short-term memory (LSTM)~\cite{LSTM}, which is a classical and famous time series model, as our baseline downstream model. 
LSTM is bidirectional with two layers and its output vectors are fed into a two-layer linear classifier with PReLU as its activation function to get the final prediction results. 
Seeing that we need to compute the difference between prediction results and the ground truth, we just adopt the cross entropy function as the loss function:
\begin{equation}
\mathcal{L}_{TFE-GNN} = \operatorname{CE}(\operatorname{Classifier}(\operatorname{LSTM}(\mathbf{z}_1, \mathbf{z}_2, \cdots, \mathbf{z}_n)), y)
\end{equation}
where $n$ is the segment length, $y$ is the ground truth and $\operatorname{CE}(\cdot)$ denotes the cross entropy function. 

Specially, we also attempt to employ a transformer layer~\cite{Self-Attention} as a downstream model, which is also an effective time series model based on the self-attention mechanism. The experimental results for transformers are also presented in the experiment section.

\section{Experiments}
\label{sec:exp}

In this section, we first present experimental settings. 
Then, we conduct experiments on multiple datasets and baselines and analyze the results. 
We also conduct an ablation study to show the effectiveness of each component in TFE-GNN. 
For comprehensive analysis, we design some model variants to evaluate the scalability of TFE-GNN and compare several baselines w.r.t. their model complexity. 
Finally, we analyse the model sensitivity of TFE-GNN. 
In detail, we conduct the experiments to answer the following questions:

\noindent \textbf{RQ1}: How is the usefulness of each component  (Section~\ref{sec:exp_ablation})?

\noindent \textbf{RQ2}: Which GNN architecture performs best (Section~\ref{sec:exp_variants})?

\noindent \textbf{RQ3}: How is the complexity of the TFE-GNN model (Section~\ref{sec:exp_complexity})?

\noindent \textbf{RQ4}: To what extent can changes in hyper-parameters affect the effectiveness of TFE-GNN (Section~\ref{sec:exp_sensitivity})?

\subsection{Experimental Settings}
\label{sec:exp_setttings}

\subsubsection{Dataset}

In order to comprehensively evaluate the effectiveness of TFE-GNN, we adopt multiple datasets, i.e., ISCX VPN-nonVPN~\cite{ISCX}, ISCX Tor-nonTor~\cite{ISCX-Tor}, and self-collected WWT datasets.

ISCX VPN-nonVPN is a public traffic dataset which contains ISCX-VPN and ISCX-nonVPN datasets. 
The ISCX-VPN dataset is collected over virtual private networks (VPNs) which are used for accessing some blocked websites or services and difficult to be recognized due to the obfuscation technology. 
Conversely, the traffic in ISCX-nonVPN is regular and not collected over VPNs. 

Similarly, ISCX Tor-nonTor is a public dataset and ISCX-Tor dataset is collected over the onion router (Tor) whose traffic can be difficult to trace. Besides, ISCX-nonTor is also regular and not collected over Tor. For comparison, we use the ISCX VPN-nonVPN and ISCX Tor-nonTor datasets with six and eight user behaviour categories, respectively. We use SplitCap to obtain bidirectional flows from public datasets. Specially, due to the scarcity of flows in the ISCX-Tor dataset, we increase the training samples by dividing each flow into 60-second non-overlapping blocks in our experiments~\cite{FlowPic}. Finally, we utilize stratified sampling to sequentially partition the training and testing dataset into 9:1 for all datasets.

The WWT dataset includes fine-grained user behaviour traffic data from three social media apps (i.e., WhatsApp, WeChat and Telegram), which have twelve, nine and six user behaviour categories, respectively. Unlike the public ISCX dataset, we additionally record the start and end timestamps of each user behaviour sample for traffic segmentation.

\subsubsection{Pre-processing}

For each dataset, we define and filter out two kinds of "anomalous" samples: \textbf{(1) Empty flows or segments}: the traffic flows or segments where all packets have no payload. \textbf{(2) Overlong flows or segments}: the traffic flows or segments whose length (i.e., the number of packets) is larger than 10000. 
An empty flow or segment does not contain any payload, thus we can not construct the corresponding graph. 
In fact, such samples are generally used to establish connections between clients and servers, having little discriminating information that helps to classify. An overlong flow or segment contains too many packets and a large number of bad packets or retransmission packets may appear in it due to temporarily bad network environment or other potential reasons. In most cases, such samples introduce too much noise, so we also consider overlong flows or segments as anomalous samples and remove them. Additionally, as for each rest sample of datasets, we remove bad packets and retransmission packets within.

For each packet in a flow or segment, we first remove the ones without payload. Then we remove the Ethernet header, which only provides some irrelevant information for classification. The source and destination IP addresses, and the port numbers are all removed for the purpose of eliminating interference with sensitive information deriving from these IP addresses and port numbers.

\subsubsection{Implementation Details and Baselines}

In the stage of traffic graph construction, we set the max packet number of one sample to 50. The max payload byte length and the max header byte length are set to 150 and 40, respectively. The PMI window size is set to 5 by default. In the stage of training, we set the max training epoch to 120. The initial learning rate is set to 1e-2 and we use the Adam optimizer with a learning rate scheduler, which gradually decays the learning rate from 1e-2 to 1e-4. The batch size is 
512, the ratio of warmup is 0.1 and the dropout rate is 0.2. We implement all models with PyTorch and run each experiment 10 times independently to take average on a single NVIDIA RTX 3080 GPU.

To give a fair comparison, 
we use four metrics, i.e., Overall Accuracy (AC), Precision (PR), Recall (RC) and Macro F1-score (F1), to evaluate TFE-GNN with following state-of-the-art baselines, including 
\textbf{\textit{Traditional Feature Engineering Based Methods}} (i.e., AppScanner~\cite{AppScanner}, CUMUL~\cite{CUMUL}, K-FP (K-Fingerprinting)~\cite{KFP}, FlowPrint~\cite{FlowPrint}, GRAIN~\cite{GRAIN}, FAAR~\cite{FAAR}, ETC-PS~\cite{ETC-PS}),
\textbf{\textit{Deep Learning Based Methods}} (i.e., FS-Net~\cite{FSNet}, EDC~\cite{EDC}, FFB~\cite{FFB}, MVML~\cite{MVML}, DF~\cite{DF}, ET-BERT~\cite{ETBERT}), and 
\textbf{\textit{Graph Neural Network Based Methods}} (i.e., GraphDApp~\cite{GraphDApp}, ECD-GNN~\cite{ECD-GNN}).

\subsection{Comparison Experiments}
\label{sec:exp_results}

\begin{table*}[t]
  \footnotesize
  \caption{Experimental Results on Self-collected WeChat, WhatsApp and Telegram Datasets}
  \label{tab:WWTresults}
  \begin{tabular}{c|cccc|cccc|cccc}
    \toprule
    Dataset & \multicolumn{4}{c|}{WeChat} & \multicolumn{4}{c|}{WhatsApp} & \multicolumn{4}{c}{Telegram} \\
    \midrule
    Model & AC & PR & RC & F1 & AC & PR & RC & F1 & AC & PR & RC & F1 \\
    \midrule
    AppScanner~\cite{AppScanner}
    & \underline{0.9927} & \underline{0.9908} & \underline{0.9904} & \underline{0.9905}
    & 0.9790 & 0.9688 & 0.9601 & 0.9628
    & 0.8379 & 0.8154 & 0.8653 & 0.8304 \\
    K-FP~\cite{KFP}
    & 0.9741 & 0.9665 & 0.9630 & 0.9645
    & 0.9710 & 0.9589 & 0.9515 & 0.9526
    & \underline{0.8797} & 0.8378 & \underline{0.8990} & \underline{0.8567} \\
    FlowPrint~\cite{FlowPrint}
    & 0.7429 & 0.5302 & 0.6380 & 0.5532
    & 0.6197 & 0.3820 & 0.4934 & 0.3880
    & 0.4833 & 0.4025 & 0.5000 & 0.4311 \\
    CUMUL~\cite{CUMUL}
    & 0.9472 & 0.9497 & 0.9412 & 0.9390
    & 0.9855 & 0.9835 & 0.9699 & 0.9756
    & 0.8330 & 0.7980 & 0.8471 & 0.8053 \\
    GRAIN~\cite{GRAIN}
    & 0.9404 & 0.9366 & 0.9369 & 0.9251
    & 0.9724 & 0.9685 & 0.9475 & 0.9550
    & 0.8003 & 0.7615 & 0.7903 & 0.7407 \\
    FAAR~\cite{FAAR}
    & 0.9873 & 0.9863 & 0.9847 & 0.9854
    & 0.9790 & 0.9683 & 0.9566 & 0.9597
    & 0.8184 & 0.7884 & 0.8507 & 0.8018 \\
    ETC-PS~\cite{ETC-PS}
    & 0.9863 & 0.9864 & 0.9831 & 0.9846
    & 0.9833 & 0.9743 & 0.9685 & 0.9703
    & 0.8477 & 0.8295 & 0.8710 & 0.8382 \\
    \midrule
    FS-Net~\cite{FSNet}
    & 0.9223 & 0.9446 & 0.9196 & 0.8964
    & \underline{0.9942} & \underline{0.9949} & \underline{0.9919} & \underline{0.9933}
    & 0.8469 & 0.8161 & 0.8744 & 0.8272 \\
    DF~\cite{DF}
    & 0.9800 & 0.9757 & 0.9751 & 0.9751
    & 0.8978 & 0.7954 & 0.8406 & 0.8143
    & 0.8251 & 0.8206 & 0.8542 & 0.7960 \\
    EDC~\cite{EDC}
    & 0.9746 & 0.9653 & 0.9625 & 0.9620
    & 0.9732 & 0.9589 & 0.9526 & 0.9546
    & 0.8153 & 0.8118 & 0.8372 & 0.7896 \\
    FFB~\cite{FFB}
    & 0.9675 & 0.9691 & 0.9661 & 0.9644
    & 0.9350 & 0.8957 & 0.8767 & 0.8708
    & 0.7990 & 0.7630 & 0.8169 & 0.7802 \\
    MVML~\cite{MVML}
    & 0.9643 & 0.9519 & 0.9540 & 0.9526
    & 0.9456 & 0.9302 & 0.9009 & 0.8971
    & 0.7943 & 0.7413 & 0.7636 & 0.7340 \\
    ET-BERT~\cite{ETBERT}
    & 0.9505 & 0.9368 & 0.9285 & 0.9266
    & 0.9107 & 0.8934 & 0.8697 & 0.8439
    & 0.7679 & \underline{0.8712} & 0.7989 & 0.7858 \\
    \midrule
    GraphDApp~\cite{GraphDApp}
    & 0.8763 & 0.8368 & 0.8378 & 0.8330
    & 0.8753 & 0.7615 & 0.8060 & 0.7791
    & 0.7766 & 0.7575 & 0.7627 & 0.7227 \\
    ECD-GNN~\cite{ECD-GNN}
    & 0.9863 & 0.9847 & 0.9830 & 0.9835
    & 0.9811 & 0.9722 & 0.9647 & 0.9672
    & 0.1810 & 0.0353 & 0.1641 & 0.0528 \\
    \midrule
    \textbf{TFE-GNN}
    & \cellcolor{zhz_gray}\textbf{0.9956} & \cellcolor{zhz_gray}\textbf{0.9953} & \cellcolor{zhz_gray}\textbf{0.9939} & \cellcolor{zhz_gray}\textbf{0.9946}
    & \cellcolor{zhz_gray}\textbf{0.9971} & \cellcolor{zhz_gray}\textbf{0.9957} & \cellcolor{zhz_gray}\textbf{0.9966} & \cellcolor{zhz_gray}\textbf{0.9961}
    & \cellcolor{zhz_gray}\textbf{0.9586} & \cellcolor{zhz_gray}\textbf{0.9584} & \cellcolor{zhz_gray}\textbf{0.9742} & \cellcolor{zhz_gray}\textbf{0.9649} \\
    \bottomrule
  \end{tabular}
\end{table*}

\begin{table*}[t]
  \caption{Experimental Results on Public ISCX VPN-nonVPN and ISCX Tor-nonTor Datasets}
  \label{tab:ISCXresults}
  \resizebox{\linewidth}{!}{
  \begin{tabular}{c|cccc|cccc|cccc|cccc}
    \toprule
    Dataset & \multicolumn{4}{c|}{ISCX-VPN} & \multicolumn{4}{c|}{ISCX-nonVPN} & \multicolumn{4}{c|}{ISCX-Tor} & \multicolumn{4}{c}{ISCX-nonTor} \\
    \midrule
    Model & AC & PR & RC & F1 & AC & PR & RC & F1 & AC & PR & RC & F1 & AC & PR & RC & F1 \\
    \midrule
    AppScanner~\cite{AppScanner}
    & 0.8889 & 0.8679 & 0.8815 & 0.8722
    & 0.7576 & 0.7594 & 7465 & 0.7486
    & 0.7543 & 0.6629 & 0.6042 & 0.6163
    & 0.9153 & 0.8435 & 0.8140 & 0.8273 \\
    K-FP~\cite{KFP}
    & 0.8713 & 0.8750 & 0.8748 & 0.8747
    & 0.7551 & 0.7478 & 0.7354 & 0.7387
    & 0.7771 & 0.7417 & 0.6209 & 0.6313
    & 0.8741 & 0.8653 & 0.7792 & 0.8167 \\
    FlowPrint~\cite{FlowPrint}
    & 0.8538 & 0.7451 & 0.7917 & 0.7566
    & 0.6944 & 0.7073 & 0.7310 & 0.7131
    & 0.2400 & 0.0300 & 0.1250 & 0.0484 
    & 0.5243 & 0.7590 & 0.6074 & 0.6153 \\
    CUMUL~\cite{CUMUL}
    & 0.7661 & 0.7531 & 0.7852 & 0.7644
    & 0.6187 & 0.5941 & 0.5971 & 0.5897
    & 0.6686 & 0.5349 & 0.4899 & 0.4997 
    & 0.8605 & 0.8143 & 0.7393 & 0.7627 \\
    GRAIN~\cite{GRAIN}
    & 0.8129 & 0.8077 & 0.8109 & 0.8027
    & 0.6667 & 0.6532 & 0.6664 & 0.6567
    & 0.6914 & 0.5253 & 0.5346 & 0.5234 
    & 0.7895 & 0.6714 & 0.6615 & 0.6613 \\
    FAAR~\cite{FAAR}
    & 0.8363 & 0.8224 & 0.8404 & 0.8291
    & 0.7374 & 0.7509 & 0.7121 & 0.7252
    & 0.6971 & 0.5915 & 0.4876 & 0.4814 
    & 0.9103 & 0.8253 & 0.7755 & 0.7959 \\
    ETC-PS~\cite{ETC-PS}
    & 0.8889 & 0.8803 & 0.8937 & 0.8851
    & 0.7273 & 0.7414 & 0.7133 & 0.7208
    & 0.7486 & 0.6811 & 0.5929 & 0.6033
    & \underline{0.9365} & \underline{0.8700} & \underline{0.8311} & \underline{0.8486} \\
    \midrule
    FS-Net~\cite{FSNet}
    & 0.9298 & 0.9263 & 0.9211 & 0.9234
    & 0.7626 & 0.7685 & 0.7534 & 0.7555
    & 0.8286 & 0.7487 & 0.7197 & 0.7242 
    & 0.9278 & 0.8368 & 0.8254 & 0.8285 \\
    DF~\cite{DF}
    & 0.8012 & 0.7799 & 0.8152 & 0.7921
    & 0.6742 & 0.6857 & 0.6717 & 0.6701
    & 0.6514 & 0.4803 & 0.4767 & 0.4719
    & 0.8568 & 0.8003 & 0.7415 & 0.7590 \\
    EDC~\cite{EDC}
    & 0.7836 & 0.7747 & 0.8108 & 0.7888
    & 0.6970 & 0.7153 & 0.7000 & 0.6978
    & 0.6400 & 0.4980 & 0.4528 & 0.4504 
    & 0.8692 & 0.7994 & 0.7411 & 0.7451 \\
    FFB~\cite{FFB}
    & 0.8304 & 0.8714 & 0.8149 & 0.8335
    & 0.7020 & 0.7274 & 0.6945 & 0.7050
    & 0.6343 & 0.4870 & 0.5203 & 0.4952 
    & 0.8954 & 0.7545 & 0.7430 & 0.7430 \\
    MVML~\cite{MVML}
    & 0.6491 & 0.7231 & 0.6198 & 0.6151
    & 0.5126 & 0.5751 & 0.4707 & 0.4806
    & 0.6343 & 0.3914 & 0.4104 & 0.3752 
    & 0.7235 & 0.5488 & 0.5512 & 0.5457 \\
    ET-BERT~\cite{ETBERT}
    & \underline{0.9532} & \underline{0.9436} & \underline{0.9507} & \underline{0.9463}
    & \cellcolor{zhz_gray}\textbf{0.9167} & \underline{0.9245} & \cellcolor{zhz_gray}\textbf{0.9229} & \underline{0.9235}
    & \underline{0.9543} & \underline{0.9242} & \underline{0.9606} & \underline{0.9397}
    & 0.9029 & 0.8560 & 0.8217 & 0.8332 \\
    \midrule
    GraphDApp~\cite{GraphDApp}
    & 0.6491 & 0.5668 & 0.6103 & 0.5740
    & 0.4495 & 0.4230 & 0.3647 & 0.3614
    & 0.4286 & 0.2557 & 0.2509 & 0.2281
    & 0.6936 & 0.5447 & 0.5398 & 0.5352 \\
    ECD-GNN~\cite{ECD-GNN}
    & 0.1111 & 0.0185 & 0.1667 & 0.0333
    & 0.0606 & 0.0101 & 0.1667 & 0.0190
    & 0.0571 & 0.0071 & 0.1250 & 0.0135 
    & 0.9078 & 0.8015 & 0.8168 & 0.7977 \\
    \midrule
    \textbf{TFE-GNN}
    & \cellcolor{zhz_gray}\textbf{0.9591} & \cellcolor{zhz_gray}\textbf{0.9526} & \cellcolor{zhz_gray}\textbf{0.9593} & \cellcolor{zhz_gray}\textbf{0.9536}
    & \underline{0.9040} & \cellcolor{zhz_gray}\textbf{0.9316} & \underline{0.9190} & \cellcolor{zhz_gray}\textbf{0.9240}
    & \cellcolor{zhz_gray}\textbf{0.9886} & \cellcolor{zhz_gray}\textbf{0.9792} & \cellcolor{zhz_gray}\textbf{0.9939} & \cellcolor{zhz_gray}\textbf{0.9855} 
    & \cellcolor{zhz_gray}\textbf{0.9390} & \cellcolor{zhz_gray}\textbf{0.8742} & \cellcolor{zhz_gray}\textbf{0.8335} & \cellcolor{zhz_gray}\textbf{0.8507} \\
    \bottomrule
  \end{tabular}
  }
\end{table*}

The comparison results on WWT and ISCX datasets are shown in Tables \ref{tab:WWTresults} and \ref{tab:ISCXresults}. 
According to Tables \ref{tab:WWTresults} and \ref{tab:ISCXresults}, we can draw the following conclusions: (1) TFE-GNN reaches the best performance compared with several baselines on the WWT dataset. Additionally, TFE-GNN also achieves the best results on four metrics, which further comprehensively demonstrates the effectiveness of our method. (2) Notably we can find that almost all the baselines perform poor on the Telegram dataset, it is due to the fact that the usage of VPNs increases the classification difficulty and introduces some background noise in case of provisionally bad network conditions caused by VPNs. However, TFE-GNN also has an outstanding result on the Telegram dataset (10.82\% f1-score improvement over the second highest), which benefits from the powerful byte encoding capability of TFE-GNN. (3) Compared with the two GNN-based methods similar to ours, i.e., GraphDApp and ECD-GNN, TFE-GNN outperforms both in all aspects. As for GraphDApp, its scheme of traffic interaction graph construction limits the expressiveness of the model. Although graphs from different traffic flows are slightly distinct in the aspect of traffic bursts, the edges between different bursts are unreasonable, which hinders feature extraction. Furthermore, an earlier burst can not "interact" with the later one using shallow GNNs because of the long and continuous connections between bursts. However, ECD-GNN is very unstable on different datasets. The reason is that the constructed graphs are lack of graph topology specificity and have highly similar structures, which significantly decreases its performance stability. With our elaborated byte-level graph construction approach, TFE-GNN can encode raw bytes well and has greater distinguishability among different traffic categories. (4) The extensive experiments on public datasets, i.e., the ISCX VPN-nonVPN and the ISCX Tor-nonTor, show that TFE-GNN can also perform well on more complicated datasets. From the Table \ref{tab:ISCXresults}, TFE-GNN is superior to almost all the baselines on public datasets except for the ISCX-nonVPN dataset, on which TFE-GNN and ET-BERT all reach similar results. However, ET-BERT is a large model with very complex model architecture while TFE-GNN is a slighter model which achieves the steadiest and best results on all datasets. We will analyse the model complexity in Section \ref{sec:exp_complexity} later.

\subsection{Ablation Study (RQ1)}
\label{sec:exp_ablation}

In this section, we conduct an ablation study of TFE-GNN on the ISCX-VPN and the ISCX-Tor datasets and show experimental results in Table \ref{tab:Ablation_study}. To facilitate the presentation of results, we denote header, payload, dual embedding module, jumping knowledge network-like concatenation, cross-gated feature fusion and activation function and batch normalization as 'H', 'P', 'DUAL', 'JKN', 'CGFF' and 'A\&N', respectively. Specially, we not only verify the effectiveness of each component in TFE-GNN, but also test the impact of some alternative modules or operations, including 'SUM' and 'MAX' operation on node features to get graph representation vectors instead of the default 'MEAN', and 'GRU' or 'TRANSFORMER' modules to serve as downstream models instead of LSTM. 

From the component ablation study of Table \ref{tab:Ablation_study}, we can draw the following conclusions: (1) The packet headers play a more important role in classification than the packet payloads and different datasets have different levels of the header and payload importance (the f1-score decreases by 2.5\% when switching the header to payload on the ISCX-VPN dataset and by 21.06\% on the ISCX-Tor dataset). (2) The usage of dual embedding increases the f1-score by 3.63\% and 0.95\%, which indicates its general effectiveness. JKN-like concatenation and cross-gated feature fusion both enhance the performance of TFE-GNN by a similar margin on two datasets. (3) We further verify the impact of the activation function and batch normalization and a significant performance drop can be seen on both datasets, which demonstrates the necessity of this two operations.

\begin{table*}[t]
  \footnotesize
  \caption{Ablation Study of TFE-GNN on ISCX-VPN and ISCX-Tor Datasets}
  \label{tab:Ablation_study}
  \begin{tabular}{c|cccccc|cccc}
    \toprule
    Method & H & P & DUAL & JKN & CGFF & A\&N & AC & PR & RC & F1 \\
    \midrule
    w/o P
    & $\checkmark$ & $\times$ & $\times$ & $\checkmark$ & $\times$ & $\checkmark$
    & 0.8713 | \underline{0.9874} & 0.8261 | \underline{0.9788} & 0.8228 | \underline{0.9934} & 0.8230 | 0.9806 \\
    w/o H
    & $\times$ & $\checkmark$ & $\times$ & $\checkmark$ & $\times$ & $\checkmark$
    & 0.8051 | 0.8726 & 0.8232 | 0.7780 & 0.7957 | 0.7809 & 0.7980 | 0.7700 \\
    w/o DUAL
    & $\checkmark$ & $\checkmark$ & $\times$ & $\checkmark$ & $\checkmark$ & $\checkmark$
    & 0.9310 | 0.9829 & 0.9149 | 0.9747 & 0.9209 | 0.9801 & 0.9173 | 0.9760 \\
    w/o JKN
    & $\checkmark$ & $\checkmark$ & $\checkmark$ & $\times$ & $\checkmark$ & $\checkmark$
    & \underline{0.9474} | 0.9790 & \underline{0.9365} | 0.9736 & \underline{0.9397} | 0.9879 & \underline{0.9374} | 0.9795 \\
    w/o CGFF
    & $\checkmark$ & $\checkmark$ & $\checkmark$ & $\checkmark$ & $\times$ & $\checkmark$
    & 0.9445 | 0.9800 & 0.9329 | 0.9717 & 0.9371 | 0.9847 & 0.9339 | 0.9770 \\
    w/o A\&N
    & $\checkmark$ & $\checkmark$ & $\checkmark$ & $\checkmark$ & $\checkmark$ & $\times$
    & 0.6105 | 0.2212 & 0.5576 | 0.0555 & 0.5487 | 0.1180 & 0.5289 | 0.0548 \\
    \midrule
    w/ SUM
    & $\checkmark$ & $\checkmark$ & $\checkmark$ & $\checkmark$ & $\checkmark$ & $\checkmark$
    & 0.8497 | 0.8194 & 0.8549 | 0.7287 & 0.8380 | 0.6986 & 0.8426 | 0.6891 \\
    w/ MAX
    & $\checkmark$ & $\checkmark$ & $\checkmark$ & $\checkmark$ & $\checkmark$ & $\checkmark$
    & 0.8480 | 0.9870 & 0.8328 | 0.9752 & 0.8115 | 0.9778 & 0.8094 | 0.9751 \\
    w/ GRU
    & $\checkmark$ & $\checkmark$ & $\checkmark$ & $\checkmark$ & $\checkmark$ & $\checkmark$
    & 0.8550 | 0.8932 & 0.8489 | 0.8702 & 0.8287 | 0.8664 & 0.8294 | 0.8610 \\
    w/ TRANSFORMER
    & $\checkmark$ & $\checkmark$ & $\checkmark$ & $\checkmark$ & $\checkmark$ & $\checkmark$
    & 0.6754 | 0.9777 & 0.5706 | 0.9753 & 0.5992 | 0.9820 & 0.5658 | \underline{0.9828} \\
    \midrule
    \textbf{TFE-GNN (default)}
    & $\checkmark$ & $\checkmark$ & $\checkmark$ & $\checkmark$ & $\checkmark$ & $\checkmark$
    & \cellcolor{zhz_gray}\textbf{0.9591 | 0.9886} & \cellcolor{zhz_gray}\textbf{0.9526 | 0.9792} & \cellcolor{zhz_gray}\textbf{0.9593 | 0.9939} & \cellcolor{zhz_gray}\textbf{0.9536 | 0.9855} \\
    \bottomrule
  \end{tabular}
\end{table*}

While on the rest part of Table \ref{tab:Ablation_study}, we can also obtain the following several points: (1) The element-wise summation on node features performs worse than the mean operation by a margin of 11.1\% and 29.64\%, respectively on two datasets w.r.t. f1-score. However, the element-wise maximum decreases the f1-score to worse results on the ISCX-VPN dataset while only decreases f1-score a little on the ISCX-Tor dataset. (2) We change the default downstream model LSTM to GRU, which worsens all metrics about \textasciitilde10\% on both datasets because of the simpler architecture of GRU. Furthermore, we employ a transformer as a downstream model for comprehensive experiments. The results show that transformer performs well on the ISCX-Tor dataset (drops f1-score within \textasciitilde1\%) while receives almost \textasciitilde40\% drop on the ISCX-VPN dataset.

\subsection{GNN Architecture Variants Study (RQ2)}
\label{sec:exp_variants}

To illustrate the scalability of GNN-based temporal fusion encoder, we select some classical GNN architectures as variants (e.g., GAT~\cite{GAT}, GIN~\cite{GIN}, GCN~\cite{GCN} and SGC~\cite{SGC}) for 
comparison on Telegram, ISCX-VPN and ISCX-Tor datasets.

\begin{figure}[!ht]
	\centering
	\includegraphics[width=0.75\linewidth]{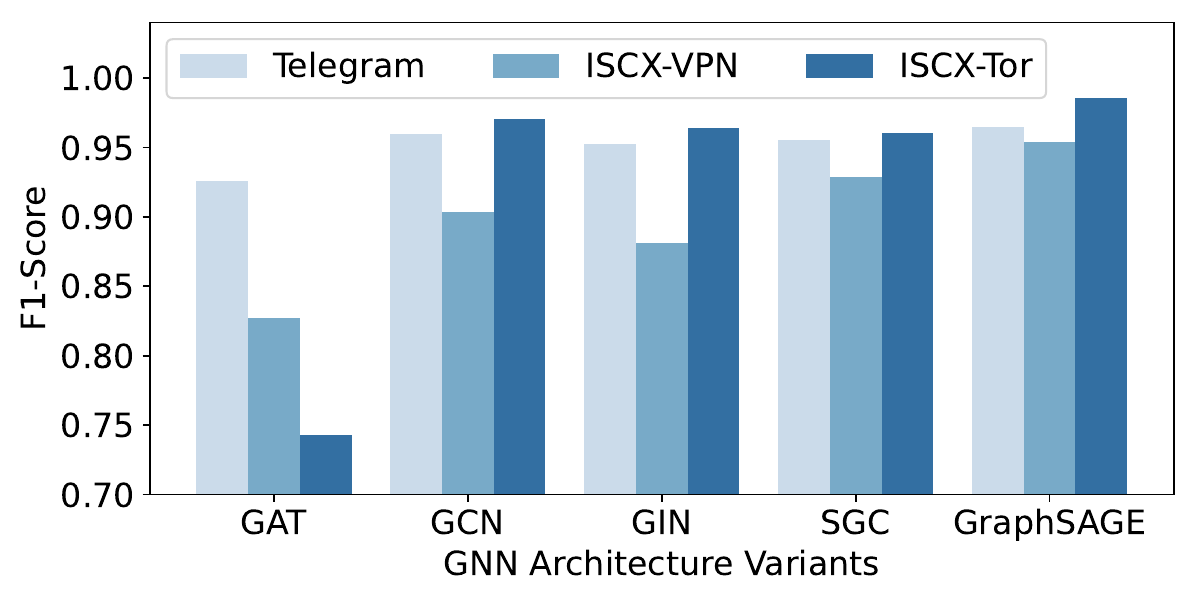}
	\caption{GNN Architecture Variants Study w.r.t. F1-score}
	\label{gnn_variants_study}
        \Description[GNN Architecture Variants Study]{We study the variants of GNN (GAT, GCN, GIN, SGC, and GraphSAGE).}
\end{figure}

From Figure \ref{gnn_variants_study}, we can find that GraphSAGE~\cite{GraphSAGE} achieves the best f1-score on three datasets. As for the rest variants, a noticeable drop in performance can be discovered, especially for GAT~\cite{GAT}. The rationale behind the results is that GNN models are easy to overfit on small-scale graphs like ours (number of nodes is up to 256). As for GAT~\cite{GAT}, the application of the attention mechanism in neighborhood feature aggregation exacerbates overfitting, which leads to a significant decline in f1-score. Among the three datasets, the relatively small fluctuation of the results on the Telegram dataset further validates the analysis above, which benefits from its larger number of training samples.

Also, a segment-level global feature can be added and shared with all nodes within one traffic graph using our model architecture if needed, which naturally takes local (packet) and global (segment) context into account simultaneously. It can be easily realized by performing concatenation, element-wise addition or other similar operations due to the strong scalability of TFE-GNN.

\subsection{Model Complexity Analysis (RQ3)}
\label{sec:exp_complexity}

To comprehensively evaluate the trade-off between model performance and model complexity, we present the floating point operations (FLOPs) and the model size of all baselines except for the traditional models in Table \ref{tab:ComplexityAnalysis}. 

From Tables \ref{tab:ComplexityAnalysis} and \ref{tab:ISCXresults}, we can draw a conclusion that TFE-GNN achieves the most significant improvement on public datasets with relatively slight model complexity increasing. Although ET-BERT reaches comparable results on the ISCX-nonVPN dataset, the FLOPs of ET-BERT are approximately five times as large as that of TFE-GNN and the number of model parameters are also doubled, which generally indicates longer model inference time and requires more computation resources. Furthermore, the pre-training stage of ET-BERT is very time-consuming and costs a lot due to the large amount of extra data during pre-training and the high model complexity. In comparison, TFE-GNN can achieve higher accuracy while reducing the training or inference costs.

\begin{table}[H]
  \footnotesize
  \caption{Model FLOPs and Parameters}
  \label{tab:ComplexityAnalysis}
  \begin{tabular}{c|c|c}
    \toprule
    Model & FLOPs(M) & Parameters(M) \\
    \midrule
    FS-Net\cite{FSNet}
    & 1.0e+2 & 3.2e+0 \\
    DF\cite{DF}
    & 2.8e+0 & 9.3e-1 \\
    EDC\cite{EDC}
    & 2.2e+1 & 2.2e+1 \\
    FFB\cite{FFB}
    & 2.6e+2 & 1.7e+0 \\
    MVML\cite{MVML}
    & 7.2e-4 & 3.7e-4 \\
    ET-BERT\cite{ETBERT}
    & 1.1e+4 & 8.6e+1 \\
    GraphDApp\cite{GraphDApp}
    & 3.8e-2 & 1.1e-2 \\
    ECD-GNN\cite{ECD-GNN}
    & 2.9e+1 & 1.4e+0 \\
    \midrule
    \textbf{TFE-GNN}
    & 2.2e+3 & 4.4e+1 \\
    \bottomrule
  \end{tabular}
\end{table}

\subsection{Model Sensitivity Analysis (RQ4)}
\label{sec:exp_sensitivity}

\begin{figure}[H]
    \centering
	\begin{subfigure}{0.325\linewidth}
		\centering
		\includegraphics[width=1.0\linewidth]{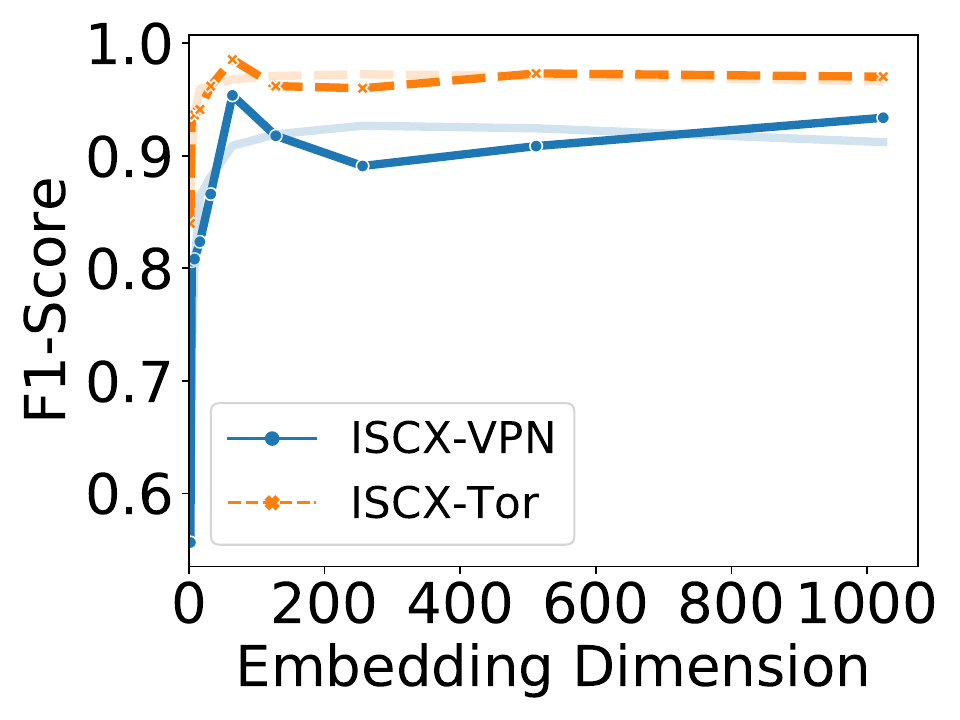}
		\caption{Embedding Dim}
		\label{embed}
	\end{subfigure}
	\centering
	\begin{subfigure}{0.325\linewidth}
		\centering
		\includegraphics[width=1.0\linewidth]{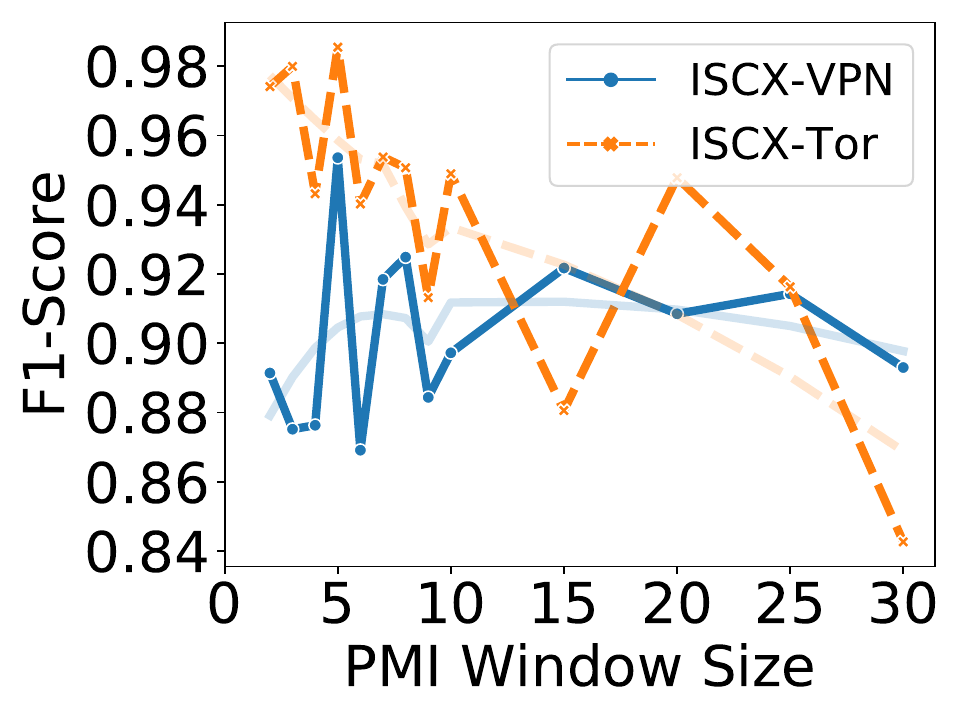}
		\caption{PMI Window Size}
		\label{w_size}
	\end{subfigure}
	\centering
	\begin{subfigure}{0.325\linewidth}
		\centering
		\includegraphics[width=1.0\linewidth]{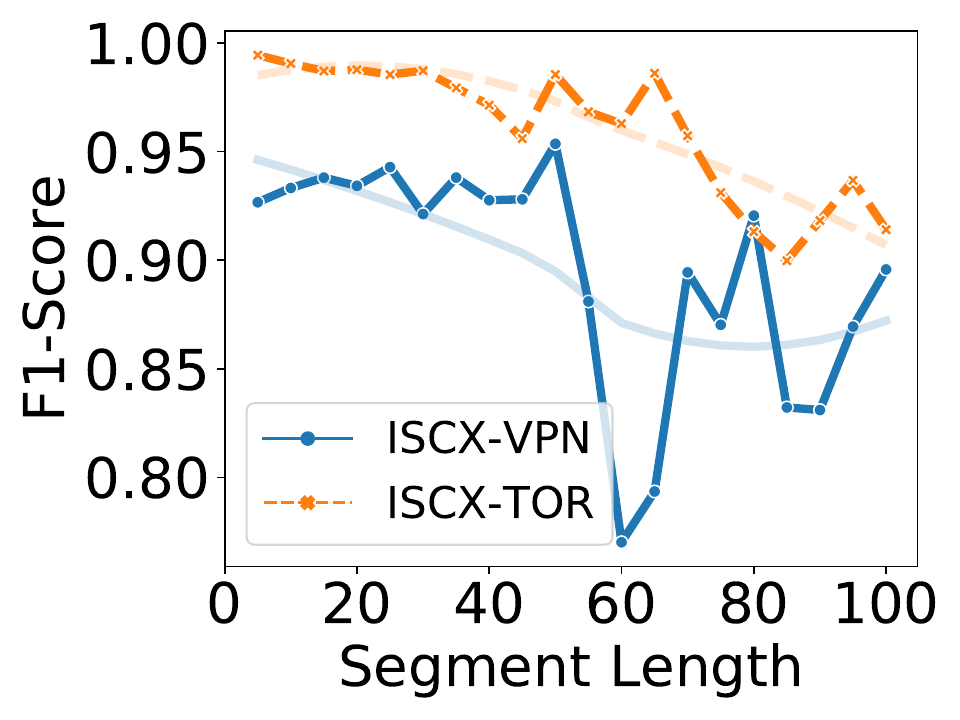}
		\caption{Segment Length}
		\label{fl}
	\end{subfigure}
	\caption{Model Sensitivity Analysis w.r.t. Dual Embedding Dimension, PMI Window Size and Segment Length on ISCX-VPN and ISCX-Tor Datasets (The shallow lines are attained by smoothing the dark plotted lines)}
	\label{sensitivity}
        \Description[Model Sensitivity Analysis]{We analyse some hyper-parameters of TFE-GNN (Embedding Dim, PMI Window Size, and Segment Length).}
\end{figure}

\textbf{(1) The Impact of Dual Embedding Dimension}. To investigate the influence of hidden dimension of the dual embedding layer, we conduct sensitivity experiments and show results in Figure \ref{embed}. As we can see, f1-score is increasing rapidly when embedding dimension is lower than 100. After that point, the model performance tends to be stable as the dimension changes. For reducing computation consuming, we just take embedding dimension 50 as our default setting. \textbf{(2) The Impact of PMI Window Size}. From Figure \ref{w_size}, we can find that a smaller window size usually results in better f1-score. The larger the window size, the more edges will be added in the traffic graphs, and the model will be harder to discriminate different traffic categories due to the too dense graphs. \textbf{(3) The Impact of Segment Length}. From Figure \ref{fl}, we can draw a conclusion that a short segment length for training usually makes the performance better. When the segment length becomes longer, more noise will be introduced and the downstream model LSTM has shortcomings in long sequence modeling, affecting the evaluation results. On the other hand, our method can achieve high accuracy when facing a short traffic flow or segment, reducing the amount of computation while improving performance.

\section{Related Work}
\label{sec:related_work}

\textbf{Traditional Feature Engineering Based Methods.}
Various methods leverage statistical features to depict the packet property and employ traditional machine learning models to conduct classification.
AppScanner uses bidirectional flow characteristics (i.e., outgoing and incoming) to extract features from traffic flows w.r.t. packet length and interval time~\cite{AppScanner}. 
CUMUL uses the features of cumulative packet length~\cite{CUMUL} and GRAIN~\cite{GRAIN} uses payload length as its features. 
ETC-PS utilizes the path signature theory to enhance the original packet length features~\cite{ETC-PS}, and Liu \emph{et al}. exploited packet length sequences using wavelet decomposition~\cite{FAAR}. 
Conti \emph{et al}.~\cite{Conti} adopts hierarchical clustering for feature extraction. 
The fingerprinting matching is also used in the traffic classification task. 
FlowPrint~\cite{FlowPrint} constructs correlation graphs as traffic fingerprinting by computing activity value between destinations IP. 
K-FP~\cite{KFP} creates fingerprinting using random forest and matches unseen samples by k-nearest neighbor. 
All of these methods suffer from the unreliable features (mentioned in Section \ref{sec:intro}).

\textbf{Deep Learning Based Methods.}
With the popularity of deep learning models, many traffic classification approaches are developed based on them. 
EDC~\cite{EDC} uses some header information of packets (e.g., protocol types, packet length and time duration) to build features for multilayer perceptions (MLPs). 
MVML~\cite{MVML} designs local and global features using packet length and time delay sequences, and simply employs a fully-connected layer for classification. Furthermore, FS-Net~\cite{FSNet}, DF~\cite{DF} as well as RBRN~\cite{RBRN} all utilize traffic flow sequences like packet length sequences to serve as the inputs of deep learning models. Additionally, DF and RBRN use convolutional neural networks (CNNs) while FS-Net utilizes gated recurrent units (GRUs) to extract temporal information of such sequences. 
For some other methods, packet bytes are used as model inputs to extract features. FFB~\cite{FFB} uses raw bytes and packet length sequences as features to feed into CNNs and RNNs. While Deep Packet~\cite{DeepPacket} utilizes CNNs and autoencoders for feature extraction. 
Recently, pre-training models are utilized to pre-train on large-scale traffic data. To give an example, ET-BERT~\cite{ETBERT} designs two novel pre-training tasks for traffic classification, which enhance the representation ability of raw bytes but are very time-consuming and costly. 
In a word, these methods can not obtain the discriminative information which is contained in raw bytes very well in a relatively efficient way, while our approach solves this pain and difficulty by introducing byte-level traffic graphs.

\textbf{Graph Neural Network Based Methods.}
Graph neural networks have strong potential in processing unstructured data and can be migrated to many fields. 
For encrypted traffic classification, GraphDApp~\cite{GraphDApp} constructs traffic interaction graphs using traffic bursts and employs graph isomorphism network~\cite{GIN} to learn representations. 
MAppGraph~\cite{MAppGraph} constructs traffic graphs based on different flows and time slices within a traffic chunk, which is almost impossible to construct a complete graph in the face of a short traffic segment. 
GCN-ETA~\cite{GCN-ETA} is a malicious traffic detection method.
To construct a graph, it will create an edge if two flows share common IP, which may result in a very dense graph.
MEMG~\cite{MEMG} utilizes markov chains to construct graphs from flows while GAP-WF~\cite{GAP-WF} maps a flow as a node in graphs and connects edges between flows which share the same identity of the clients. 
Besides, Huoh \emph{et al}.~\cite{ECD-GNN} directly created edges based on the chronological relationship of packets among a flow, being lack of specificity. These methods all construct graphs at the level of traffic flows, which are vulnerable if there is too much noise within flows.

\section{Conclusion and Future Work}
\label{sec:conclusion}

We propose an approach to construct byte-level traffic graphs and a model named TFE-GNN for encrypted traffic classification. 
The byte-level traffic graph construction approach can mine the potential correlation between raw bytes and generate discriminative traffic graphs. TFE-GNN is designed to extract high-dimensional features from constructed traffic graphs. Finally, TFE-GNN can encode each packet into an overall representation vector, which can be used for some downstream tasks like traffic classification. Several baselines are selected to evaluate the effectiveness of TFE-GNN. 
The experimental results show that our proposed model comprehensively surpasses all the baselines on the WWT and the ISCX datasets. 
Elaborately designed experiments further demonstrate that TFE-GNN has strong effectiveness. 

In the future, we will attempt to improve TFE-GNN in terms of the following limitations. 
\textbf{(1) Limited graph construction approach}. The graph topology of the proposed model is determined before the training procedure, which may result in non-optimal performance. Moreover, the TFE-GNN can not cope with the byte-level noise implied in the raw bytes of each packet. \textbf{(2) Unused temporal information implied in byte sequences}. The byte-level traffic graphs are constructed without introducing the explicit temporal characteristics of byte sequences.

\newpage

\begin{acks}
This work was supported in part by the National Natural Science Foundation of China (61972219,62202406,61972189), the Research and Development Program of Shenzhen (JCYJ20190813174403598, SGDX20190918101201696), the Overseas Research Cooperation Fund of Tsinghua Shenzhen International Graduate School (HW2021013), Shenzhen Science and Technology Innovation Commission: Research Center for Computer Network (Shenzhen) Ministry of Education, HK ITF Project (GHP/052/19SZ), and the Key Laboratory of Network Assessment Technology, Institute of Information Engineering, Chinese Academy of Sciences, Beijing, China.
\end{acks}

\bibliographystyle{ACM-Reference-Format}
\bibliography{ref}

\newpage

\appendix

\section{Threat Model and Assumptions}
\label{sec:appendix_threat}

In this section, we present the threat model and assumptions. Normal users employ mobile apps to communicate with remote servers. The attacker is a passive observer (i.e., he cannot decrypt or modify packets). The attacker captures the packets of the target apps by compromising the device or sniffing the network link. Then, the attacker analyzes the captured packets to infer the behaviors of normal users.

\section{Long-tailed Distribution of the ISCX Dataset}
\label{sec:appendix_long-tailed}

We count the flow length on three datasets, i.e., ISCX-VPN, ISCX-NonVPN and ISCX-NonTor datasets and the figure below shows that the flow length generally obeys the long-tailed distribution.

\begin{figure}[H]
	\centering
	\includegraphics[width=1.0\linewidth]{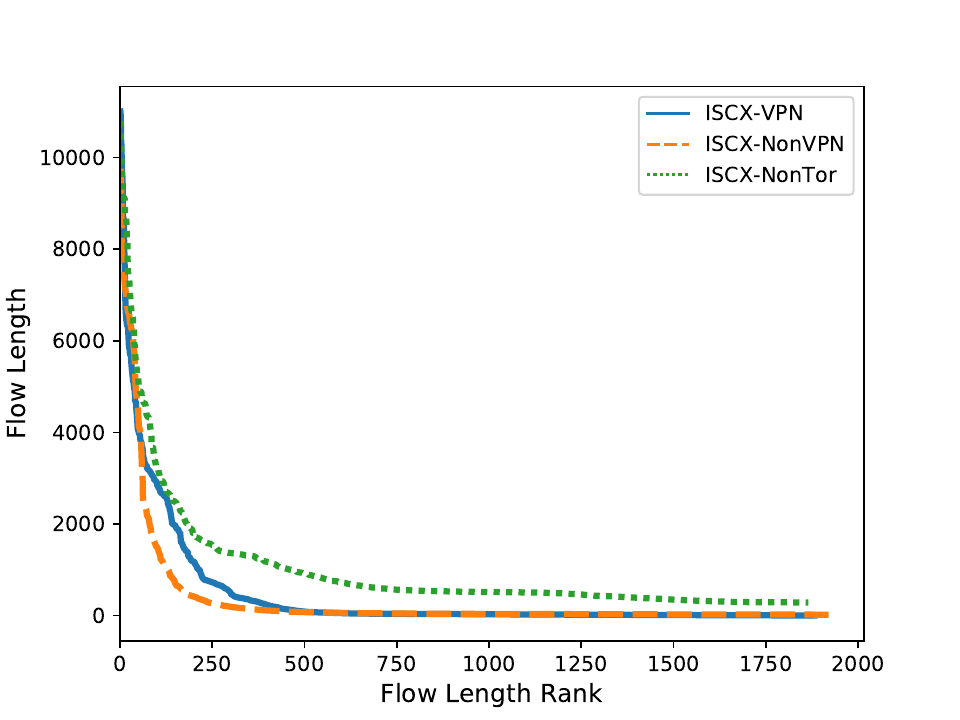}
	\caption{Long-tailed Distribution of the Flow Length on ISCX-VPN, ISCX-NonVPN and ISCX-NonTor Datasets (Considering the amount of data, only part of the sorted data is shown)}
	\label{long-tailed}
        \Description[Long-tailed Distribution]{The long-tailed distribution of traffic flow length.}
\end{figure}

\end{document}